# Device Image-IV Mapping using Variational Autoencoder for Inverse Design and Forward Prediction

Thomas Lu, Albert Lu, and Hiu Yung Wong, *Senior Member, IEEE*

*Abstract*—This paper demonstrates the learning of the underlying device physics by mapping device structure images to their corresponding Current-Voltage (IV) characteristics using a novel framework based on variational autoencoders (VAE). Since VAE is used, domain expertise is not required and the framework can be quickly deployed on any new device and measurement. This is expected to be useful in the compact modeling of novel devices when only device cross-sectional images and electrical characteristics are available (e.g. novel emerging memory). Technology Computer-Aided Design (TCAD) generated and hand-drawn Metal-Oxide-Semiconductor (MOS) device images and noisy drain-current-gate-voltage curves ($I_DV_G$) are used for the demonstration. The framework is formed by stacking two VAEs (one for image manifold learning and one for $I_DV_G$ manifold learning) which communicate with each other through the latent variables. Five independent variables with different strengths are used. It is shown that it can perform inverse design (generate a design structure for a given $I_DV_G$) and forward prediction (predict $I_DV_G$ for a given structure image, which can be used for compact modeling if the image is treated as device parameters) successfully. Since manifold learning is used, the machine is shown to be robust against noise in the inputs (i.e. using hand-drawn images and noisy $I_DV_G$ curves) and not confused by weak and irrelevant independent variables.

## I. Introduction

DUE to the ever-more powerful machine learning techniques, Technology Computer-Aided Design has been a cost-effective tool that can provide abundant data to augment machine learning (TCAD-augmented ML) in semiconductor device design automation. It has been used for defect analysis [1]-[6], device characteristic predictions [7][8], device/circuit manifold learning [9]-[11], inverse design [12], and surrogate model development [13][14].

However, to the best of our knowledge, all studies so far use either device electrical characteristics (e.g. Current-Voltage (IV) curves, Capacitance-Voltage curves (CV), threshold voltage ($V_{TH}$)) or numerical description of the principal characteristics (e.g. gate length, $L_G$) of the device structures as the inputs to ML and no study has been performed to correlate the device image to its electrical characteristics.

For any novel device which is not well understood, it is difficult to extract the principal characteristics from its structure image. It is also possible that such characteristics cannot be extracted easily (such as in the geometry of an optimized multiplexer in photonics [15]). Moreover, some parameters which are important to human eyes (e.g. gate poly-Si thickness) are weak and irrelevant independent variables to the IV. Therefore, it is important to investigate the possibility of learning the underlying device physics using images in which domain expertise is not used. If this is successful, it will create a novel approach to augment traditional compact modeling by using novel device images (instead of parameters such as $L_G$) as the input to a compact model to predict the electrical characteristics before the physics of the novel device is understood.

In this paper, we demonstrate that it is possible to train a machine to understand the underlying device physics and correlate the structure image with the electrical characteristics. We show that the machine is not confused by weak and irrelevant independent variables and is robust to noise. No domain expertise is required to preprocess the data. The machine can be used to perform inverse design and forward prediction. While a traditional MOSFET structure and its $I_DV_G$ are used as a demonstration, such an approach is expected to be useful to help understand the physics of novel devices (e.g. using enough SEM/TEM images and the corresponding IV to understand the behavior of a novel emerging memory).

## II. Data Generation and Machine Building

Planar MOSFETs are used in this study. This is because the 2D cross-section of planar MOSFETs is more complicated than the cross-section of FinFET and nanowire due to the existence of a lightly doped drain (LDD) and the lack of symmetry in the vertical direction. Therefore, it is a better test of the capability of the approach. 2D cross-section is used because 3D images

This material is based upon work supported by the National Science Foundation under Grant No. 2046220. *(Corresponding author: Hiu Yung Wong huiyung.wong@sjsu.edu).*

Thomas Lu, Albert Lu, and Hiu Yung Wong are with the M-PAC Lab, Electrical Engineering Department, San Jose State University, San Jose, CA, USA. Thomas Lu is also affiliated with Stanford Online High School, CA, USA. Albert Lu, is also affiliated with the Computer Engineering Department, San Jose State University, San Jose, CA, USA. Thomas Lu and Albert Lu have equal contribution.

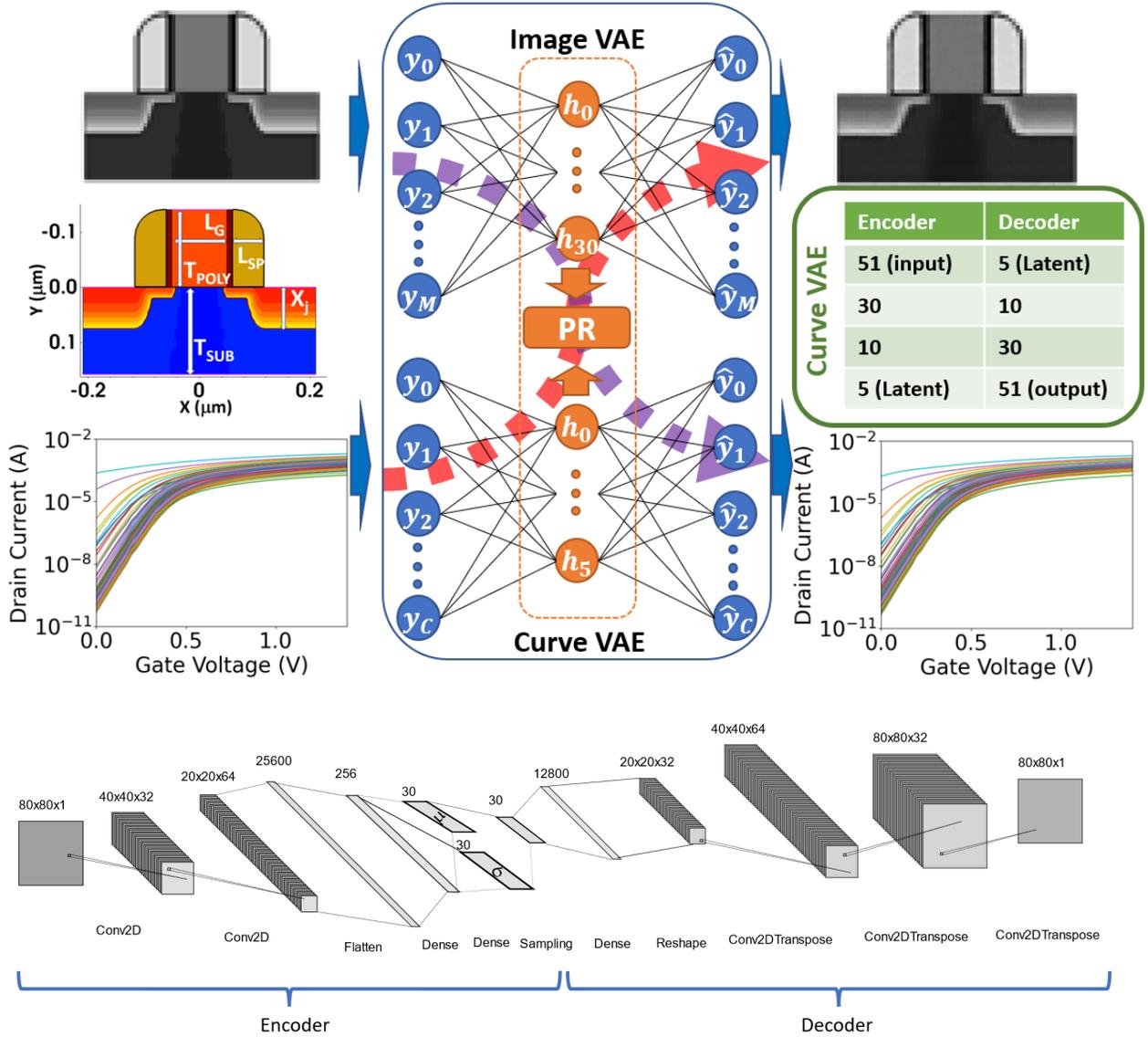

Fig. 1. Top: The machine used in this study which combines an image-VAE, a curve-VAE, and 3rd order Polynomial Regressors (PR) (enclosed in the blue box). The image- (curve-) VAE is trained with the corresponding images ($I_D V_G$ curves). Only 3 layers are shown for clarity. The PRs are trained to correlate the latent variables between the image-VAE and curve-VAE (enclosed in the orange box). Purple (red) thick dashed lines indicate the forward (inverse) prediction paths. M = 6399 and C = 50. A summary of the curve-VAE structure is shown in the figure by the number of nodes in each layer (green box). Bottom: The full structure of the image-VAE. RELU is used for activation except the last layer (sigmoid for image-VAE and linear activation for curve-VAE). One image example and a group of 100 randomly selected curves are shown as the inputs and outputs of the image- and curve-VAE, respectively in the top figure.

are generally not available in the experiment. Moreover, both well-behaved and leaky planar MOSFETs are used. This is to demonstrate that the machine can learn the complex underlying physics instead of just a simple compact model of a well-behaved transistor, thus, it has the potential to learn non-transistor physics. It should be noted that, unlike most compact models which are developed for well-behaved devices, Technology Computer-Aided-Design (TCAD) simulation is valid for both well-behaved and leaky devices as long as the appropriate models are turned on.

8755 MOSFETs (8000 for training and 755 for testing) with various gate lengths ($L_G \in [25nm, 290nm]$), source/drain junction depths ($X_j \in [10nm, 90nm]$), spacer widths ($L_{SP} \in [10nm, 110nm]$), gate poly thicknesses ($T_{POLY} \in [50nm, 150nm]$), and substrate thicknesses ($T_{SUB} \in [100nm, 200nm]$) are generated uniformly and randomly (Fig. 1) using TCAD Sentaurus [16]. From device physics, it is known that $L_G$ and $X_j$ are strong parameters affecting $I_{ON}$ and $I_{OFF}$. $L_{SP}$ has a certain strength and sometimes affects $I_{OFF}$ if $X_j/L_G$ is large. $T_{POLY}$ and $T_{SUB}$ are weak and irrelevant parameters. The weak parameters are added to show that the framework will not be confused by irrelevant independent variables. The gate oxide thickness, poly gate doping, and substrate doping are 1.4 nm, $10^{20} cm^{-3}$, and $1.5 \times 10^{18} cm^{-3}$, respectively. The transistor width is 1μm. The structures are simulated in SDevice with a standard setup including Fermi-Dirac statistics, density gradient for quantum correction, high field saturation, and IALMob [16] to produce saturation $I_D V_G$ curves with $V_G$ from 0 V to 1.4 V and $V_D$ = 1.4 V. The structure image is saved in monochromatic portable network graphic (png) format with 80×80 pixels and the $I_DV_G$ is discretized to 51 points. A minimum $L_G$ of 25nm is used because this corresponds to only 5 pixels in the figure.

Variational autoencoder-based machines are trained to perform the inverse design and forward prediction. Fig. 1 shows that the machine is composed of two major components, namely one image-VAE (top) and one curve-VAE (bottom) [17]. Unlike in a regular autoencoder (AE) [4][9] where only the difference between the input and the output (reconstruction loss) is minimized, a VAE is trained also by regularizing the encoding (data in the latent space) distribution. This is done by adding the Kulback-Leibler (KL) divergence between the encoding distribution and the Gaussian distribution to the loss function [17]. With this, the VAE is less susceptible to overfitting. Since this work is related to image feature extraction, VAE is chosen to avoid spurious image prediction due to overfitting as inspired by the work of using VAE to generate handwritten digits [17].

Since the structures are generated by varying 5 parameters ($L_G$, $X_j$, $L_{SP}$, $T_{POLY}$, and $T_{SUB}$), 5 latent nodes are enough to capture the underlying physics [3][9]. However, more latent nodes are found to help find the global minimum and improve training performance during the VAE training. Therefore, the image-VAE has 30 latent variables. In addition to KL-loss, binary cross-entropy and mean squared error loss is used as the reconstruction loss in the image- and curve-VAE, respectively. Fig. 1 shows that the trained VAEs can reconstruct the testing input image and curves (left) accurately at their outputs (right). To connect the image-VAE and curve-VAE, their latent variables, $h_i$, are regressed against each other in both directions using $3^{rd}$-order polynomial regressions (orange box in Fig. 1).

Note that 8000 data are required for accurate training because there are 5 parameters. This means that, on average, $8000^{1/5} \sim 6$ variations are required for each parameter to learn the underlying physics. If only considering the 3 medium to strong parameters ($L_G$, $X_j$, and $L_{SP}$), it is expected that only $6^3=216$ data are required. This is consistent with the finding in manifold learning using AE in [9].

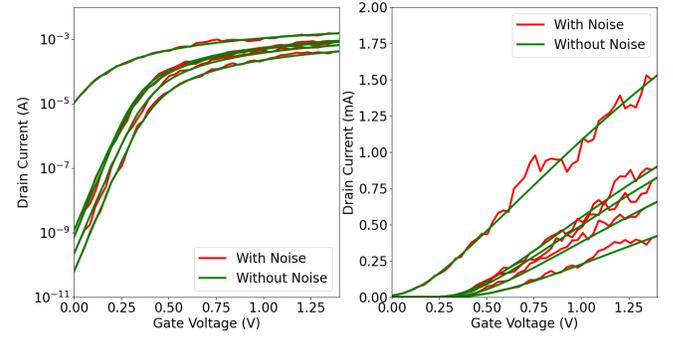

Fig. 2. Five selected $I_DV_G$ curves with and without noise used in the inverse design. Left: logarithmic scale. Right: linear scale.

## III. INVERSE DESIGN

Inverse design in engineering is generally a very difficult problem [15]. In this study, inverse design refers to the finding of a MOSFET structure to reproduce a given $I_DV_G$. *It should be noted that the goal of inverse design is not to reproduce a certain device structure but to produce a device that has a certain $I_DV_G$*. Very often, the device engineer needs to design a transistor to match the $I_{OFF}$ ($V_G = 0V$, $V_D = 1.4V$) and $I_{ON}$ ($V_G = 1.4V$, $V_D = 1.4V$) requirements by a circuit designer. This corresponds to the "red path" in Fig. 1.

To demonstrate that the framework is capable of performing inverse design and to show that the machine has learned the underlying physics instead of memorizing the training data, "hand-drawn" curves are used. The "hand-drawn" curves are constructed by adding noise to the $I_DV_G$ curves it has not seen before. To keep the $I_{OFF}$ and $I_{ON}$ unchanged (as per the specification), noise is not added to the terminal points ($V_G = 0V$, 0.028V, 1.372V, 1.4V). Twenty noisy curves are tested and Fig. 2 shows 5 random $I_DV_G$ curves from the test set with and without noise. The original structures corresponding to these curves are shown in Fig. 3 (corresponding to $L_G = 47.0$nm, 92.0 nm, 114.0 nm, 135.0 nm, and 247.0 nm, respectively). The inverse-designed structures, which are obtained by feeding the perform $I_D V_G$ simulations with $I_{ON}$ and $I_{OFF}$ extracted. They are then plotted against the $I_{ON}$ and $I_{OFF}$ of the required values. The $R^2$ is found to be 0.90 and 0.97, respectively (Fig. 4 and Table I), which shows that it can capture the physics well.

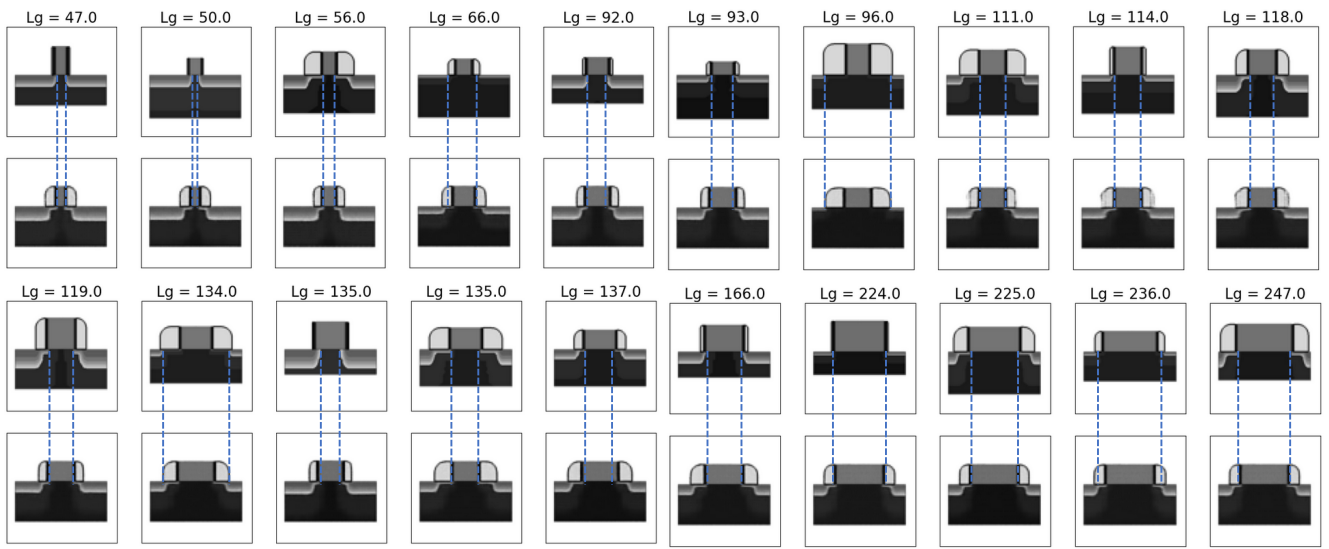

Fig. 3. Twenty structures are shown with varying $L_g$. The top of each row has the structures giving the exact $I_DV_G$ curves without noise. The bottom of each row has the inverse designed structures corresponding to the $I_DV_G$ curves with noise.

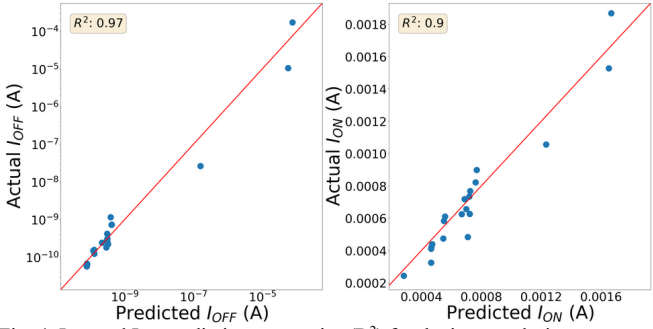

Fig. 4. $I_{OFF}$ and $I_{ON}$ prediction accuracies ($R^2$) for the inverse design.

TABLE I
$I_{ON}$ AND $I_{OFF}$ PREDICTION ACCURACIES ($R^2$)

| Machine Parameters | INVERSE DESIGN | | FORWARD PREDICTION | |
|---|---|---|---|---|
| | $I_{ON}$ | $I_{OFF}$ | $I_{ON}$ | $I_{OFF}$ |
| $R^2$ | 0.90 | 0.97 | 0.97 | 0.96 |

noisy "hand-drawn" $I_D V_G$ in Fig. 2 into the red path in Fig. 1, are shown at the bottom of each row in Fig. 3.

It is found that passing the noisy curves through the curve-VAE 2 times before the red path and passing the final images 3 times through the image-VAE after the red path improves the results.

There are a few observations. 1) The inverse-designed structures give a correct trend in effective $L_G$ (distance between LDD, guided by the blue dotted line) which determines the $I_{ON}$ (note that the spacer LDD does not contribute too much to the ON-state resistance). Since the effective $L_G$ is *not* extracted and given to the machine manually, the machine has successfully learned the underlying physics itself. 2) It is known from device physics that $T_{POLY}$ and $T_{SUB}$ do not affect $I_D V_G$. Therefore, the machine has learned to just assume a constant $T_{POLY}$ and $T_{SUB}$ for all devices. 3) For $L_G$ = 135nm, the actual device has a large $X_j$ which is expected to increase the $I_{OFF}$. However, since the substrate contact is put at the bottom and close to the S/D, this reduces the short-channel effect substantially. As a result, the machine predicts a device with a smaller $X_j$ at a larger $T_{SUB}$. As mentioned earlier, the goal of inverse design is *not* to reproduce the original structure but to generate a structure that produces the desired $I_D V_G$. Point 3) shows that this is successful.

To confirm that it has produced the desired electrical characteristics, the inverse-designed structures are then used to

## IV. FORWARD PREDICTION

It is also desirable to predict the $I_D V_G$ for a given structure without using time-consuming TCAD simulations. This can be used as a novel approach to augment compact modeling when a novel device is still not well-understood. In that case, the image can be used as the input parameters to the compact model without the need for domain expertise to extract the strong parameters (e.g. $L_G$ in this case).

This can be achieved by using the forward prediction path (purple path in Fig. 1). To demonstrate that the machine has learned the physics, hand-drawn structures are used. The hand-drawn structures are obtained by modifying the unseen test set structures. Fig. 5 shows all 10 of the test structures (top) and the corresponding hand-drawn structures (middle). Note that the hand-drawn structures are drawn so that the 5 parameters are the same as the original structure, but it has a different greyscale (e.g. spacer has different intensity and the doping gradient information of S/D and the substrate is lost) and unsmooth boundaries.

The corresponding $I_{OFF}$ and $I_{ON}$ of all curves are also extracted. They are then plotted against the expected $I_{OFF}$ and $I_{ON}$.

It is found that the prediction of both $I_{OFF}$ and $I_{ON}$ are good with $R^2$ = 0.96 and 0.97, respectively (Fig. 6 and Table I). Note that to improve the result, the hand-drawn images are passed through the image-VAE one time before the purple path.

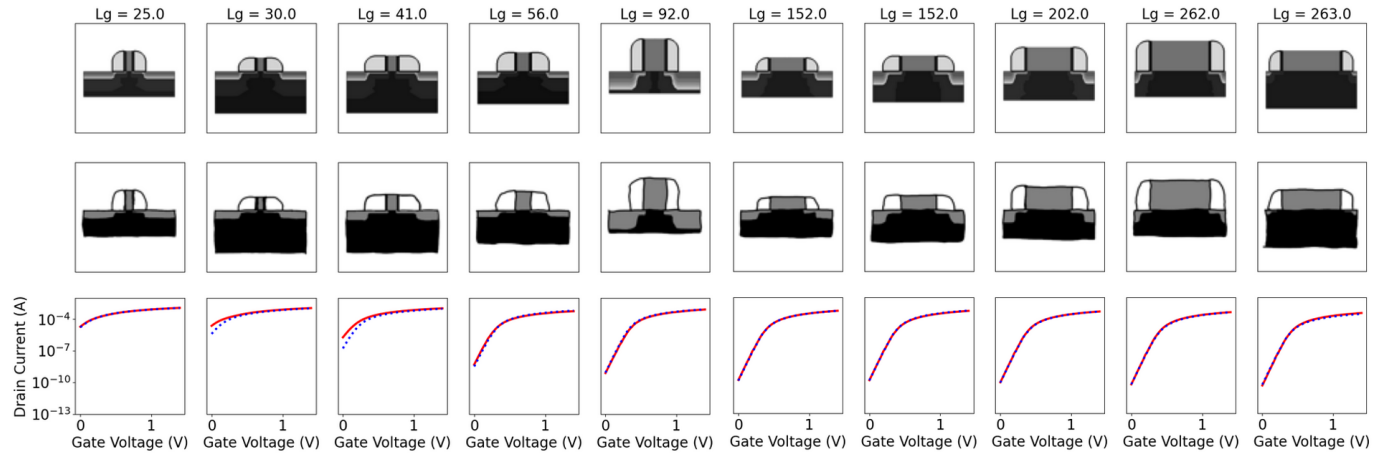

Fig. 5. Ten of the structures used to test the forward prediction process. Top: TCAD generated structures. Middle: The corresponding hand modified structures. Bottom: TCAD simulated $I_D V_G$ based on TCAD generated structure (dotted blue) and $I_D V_G$ from hand drawn structure by forward design (red line). $L_G$ unit is in nm.

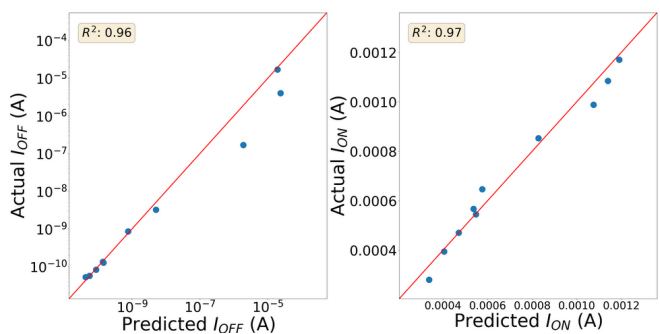

Fig. 6. $I_{OFF}$ and $I_{ON}$ prediction accuracies ($R^2$) for the forward design.

## V. Conclusions

A novel framework based on VAE is demonstrated to have effectively learned the underlying physics by mapping the device image to $I_DV_G$ curves. It shows good results in performing inverse design and forward prediction even with noisy curves and hand-drawn images. This can be used to learn the physics of novel devices when pictures are available. Particularly, the forward prediction can be used to augment the traditional compact modeling by using the image of a device as the input parameters without the need for domain expertise to extract the strong parameters. This will be very useful for novel devices which are not well-understood yet.

## References

[1] Y. S. Bankapalli and H. Y. Wong, "TCAD Augmented Machine Learning for Semiconductor Device Failure Troubleshooting and Reverse Engineering," *2019 International Conference on Simulation of Semiconductor Processes and Devices (SISPAD)*, Udine, Italy, 2019, pp. 1-4, doi: 10.1109/SISPAD.2019.8870467.

[2] C. Teo, K. L. Low, V. Narang and A. V. Thean, "TCAD-Enabled Machine Learning Defect Prediction to Accelerate Advanced Semiconductor Device Failure Analysis," *2019 International Conference on Simulation of Semiconductor Processes and Devices (SISPAD)*, Udine, Italy, 2019, pp. 1-4, doi: 10.1109/SISPAD.2019.8870440.

[3] T. Lu, V. Kanchi, K. Mehta, S. Oza, T. Ho and H. Y. Wong, "Rapid MOSFET Contact Resistance Extraction From Circuit Using SPICE-Augmented Machine Learning Without Feature Extraction," in IEEE Transactions on Electron Devices, vol. 68, no. 12, pp. 6026-6032, Dec. 2021, doi: 10.1109/TED.2021.3123092.

[4] H. Dhillon, K. Mehta, M. Xiao, B. Wang, Y. Zhang and H. Y. Wong, "TCAD-Augmented Machine Learning With and Without Domain Expertise," in IEEE Transactions on Electron Devices, vol. 68, no. 11, pp. 5498-5503, Nov. 2021, doi: 10.1109/TED.2021.3073378.

[5] S. S. Raju, B. Wang, K. Mehta, M. Xiao, Y. Zhang and H. -Y. Wong, "Application of Noise to Avoid Overfitting in TCAD Augmented Machine Learning," 2020 International Conference on Simulation of Semiconductor Processes and Devices (SISPAD), Kobe, Japan, 2020, pp. 351-354, doi: 10.23919/SISPAD49475.2020.9241654.

[6] H. Y. Wong, M. Xiao, B. Wang, Y. K. Chiu, X. Yan, J. Ma, K. Sasaki, H. Wang and Y. Zhang, "TCAD-Machine Learning Framework for Device Variation and Operating Temperature Analysis With Experimental Demonstration," in *IEEE Journal of the Electron Devices Society*, vol. 8, pp. 992-1000, 2020, doi: 10.1109/JEDS.2020.3024669.

[7] J. Chen, M. B. Alawieh, Y. Lin, M. Zhang, J. Zhang, Y. Guo and D. Z. Pan, "Powernet: SOI Lateral Power Device Breakdown Prediction With Deep Neural Networks," in *IEEE Access*, vol. 8, pp. 25372-25382, 2020, doi: 10.1109/ACCESS.2020.2970966.

[8] H. Carrillo-Nuñez, N. Dimitrova, A. Asenov and V. Georgiev, "Machine Learning Approach for Predicting the Effect of Statistical Variability in Si Junctionless Nanowire Transistors," in *IEEE Electron Device Letters*, vol. 40, no. 9, pp. 1366-1369, Sept. 2019. doi: 10.1109/LED.2019.2931839.

[9] K. Mehta and H. -Y. Wong, "Prediction of FinFET Current-Voltage and Capacitance-Voltage Curves Using Machine Learning With Autoencoder," in IEEE Electron Device Letters, vol. 42, no. 2, pp. 136-139, Feb. 2021, doi: 10.1109/LED.2020.3045064.

[10] V. Eranki, N. Yee and H. Y. Wong, "Out-of-training-range Synthetic FinFET and Inverter Data Generation using a Modified Generative Adversarial Network," in IEEE Electron Device Letters, 2022, doi: 10.1109/LED.2022.3207784.

[11] V. Eranki, T. Lu, and H. Y. Wong, "Comparison of Manifold Learning Algorithms for Rapid Circuit Defect Extraction in SPICE-Augmented Machine Learning," 2022 IEEE 19th Annual Workshop on Microelectronics and Electron Devices (WMED), 2022, pp. 1-4, doi: 10.1109/WMED55302.2022.9758032.

[12] K. Mehta, S. S. Raju, M. Xiao, B. Wang, Y. Zhang and H. Y. Wong, "Improvement of TCAD Augmented Machine Learning Using Autoencoder for Semiconductor Variation Identification and Inverse Design," in *IEEE Access*, vol. 8, pp. 143519-143529, 2020, doi: 10.1109/ACCESS.2020.3014470.

[13] R. Wang, C. Chen, Q. Huang, Y. Wang, C. Hu, D. Wu, J. Wang, and R. Huang, "New-Generation Design-Technology Co-Optimization (DTCO): Machine-Learning Assisted Modeling Framework," 2019 Silicon Nanoelectronics Workshop (SNW), Kyoto, Japan, 2019, pp. 1-2, doi: 10.23919/SNW.2019.8782897.

[14] A. Lu, J. Marshall, Y. Wang, M. Xiao, Y. Zhang, and H. Y. Wong, "Vertical GaN Diode BV Maximization through Rapid TCAD Simulation and ML-enabled Surrogate Model," Solid-State Electronics, Volume 198, December 2022, 108468, https://doi.org/10.1016/j.sse.2022.108468.

[15] A. Piggott, J. Lu, K. G. Lagoudakis, J. Petykiewicz, T. M. Babinec and J. Vuckovic, "Inverse design and demonstration of a compact and broadband on-chip wavelength demultiplexer," Nature Photon 9, 374–377 (2015). https://doi.org/10.1038/nphoton.2015.69

[16] Sentaurus™ Device User Guide Version S-2021.06, June. 2021.

[17] Diederik P Kingma and Max Welling, "Auto-encoding variational bayes," arXiv preprint arXiv:1312.6114, 2013.